\documentclass[letterpaper, 10 pt, conference]{ieeeconf}
\IEEEoverridecommandlockouts    
\usepackage{graphics}           
\usepackage{times}              
\usepackage{amsmath}            
\usepackage{amssymb}            
\usepackage{graphicx}
\usepackage{algorithm}
\usepackage[noend]{algpseudocode}
\usepackage{booktabs}
\usepackage{color}
\definecolor{instructioncolor}{rgb}{.5,.5,.5}

\usepackage[font=small]{caption}

\def\secref#1{Sec.~\ref{#1}}
\def\figref#1{Fig.~\ref{#1}}
\def\tabref#1{Tab.~\ref{#1}}
\def\eqref#1{Eq.~(\ref{#1})}

\makeatletter
\usepackage{xspace}
\DeclareRobustCommand\onedot{\futurelet\@let@token\@onedot}
\def\@onedot{\ifx\@let@token.\else.\null\fi\xspace}

\def\etal{{et al}\onedot}
\makeatother

\usepackage{array}
\newcolumntype{L}[1]{>{\raggedright\let\newline\\\arraybackslash\hspace{0pt}}m{#1}}
\newcolumntype{C}[1]{>{\centering\let\newline\\\arraybackslash\hspace{0pt}}m{#1}}
\newcolumntype{R}[1]{>{\raggedleft\let\newline\\\arraybackslash\hspace{0pt}}m{#1}}

\usepackage{multirow}
\usepackage[colorinlistoftodos,prependcaption,textsize=footnotesize]{todonotes}

\renewcommand{\baselinestretch}{0.98} 
\title{\LARGE \bf Predicting Dense and Context-aware Cost Maps\\ for Semantic Robot Navigation}

\author{Yash Goel${}^*{}^{1, 2}$, Narunas Vaskevicius${}^*{}^2$, Luigi Palmieri${}^2$, Nived Chebrolu${}^3$, Cyrill Stachniss${}^{1,4}$
\thanks{$^{1}$Y.\,Goel is with the laboratory for Photogrammetry and Robotics, University of Bonn, Germany, and Robert Bosch GmbH.
        {\tt\footnotesize  \{s7yagoel\}@uni-bonn.de.}}%
\thanks{$^{2}$L.\,Palmieri, N.\,Vaskevicius are with Robert Bosch GmbH, Corporate Research, Stuttgart, Germany.
        {\tt\footnotesize  \{luigi.palmieri, narunas.vaskevicius\}@de.bosch.com.}} %
\thanks{$^{3}$N.\,Chebrolu is with with the Dynamic Robot Systems Group, University of Oxford, UK.
        {\tt\footnotesize  \{nived\}@robots.ox.ac.}}%
\thanks{$^{1,4}$C.\,Stachniss is with the University of Bonn, Germany, with the Department of Engineering Science at the University of Oxford, UK, and with the Lamarr Institute for Machine Learning and Artificial Intelligence, Germany.
        {\tt\footnotesize  \{cyrill.stachniss\}@igg.uni-bonn.de.}}%
\thanks{$^{*}$Denotes equal contribution.}%
\thanks{This work was partly supported by the EU Horizon 2020 research and innovation program under grant agreement No. 101017274 (DARKO).}%
}

\begin{document}
\maketitle
\thispagestyle{empty}
\pagestyle{empty}

\begin{abstract}

  We investigate the task of object goal navigation in unknown environments where the target is specified by a semantic label (e.g. find a couch). Such a navigation task is especially challenging as it requires understanding of semantic context in diverse settings. Most of the prior work tackles this problem under the assumption of a discrete action policy whereas we present an approach with continuous control which brings it closer to real world applications. We propose a deep neural network architecture and loss function to predict dense cost maps that implicitly contain semantic context and guide the robot towards the semantic goal. We also present a novel way of fusing mid-level visual representations in our architecture to provide additional semantic cues for cost map prediction. The estimated cost maps are then used by a sampling-based model predictive controller (MPC) for generating continuous robot actions. The preliminary experiments suggest that the cost maps generated by our network are suitable for the MPC and can guide the agent to the semantic goal more efficiently than a baseline approach. 
  The results also indicate the importance of mid-level representations for navigation by improving the success rate by 7 percentage points.
\end{abstract}
\section{Introduction}
\label{sec:intro}

Equipping a robot with human-like semantic aware navigation skills is essential for achieving true autonomy. A robot tasked with finding a couch should be able to draw conclusion that if it is near a TV then the couch should be nearby - since they generally tend to be in the same space (\textit{living room}). It is important to learn these semantic relationships between various objects in the environment and also between different spaces in the environment. This high level understanding can then be used to navigate the robot to target objects or area.

Reliable and accurate semantic robot navigation is still an open research question \cite{crespo2020surveyn}. Traditional approaches use semantic knowledge for building graphs \cite{rogers2013icra} or try to navigate to rooms \cite{talbot2016icra} using various planning approaches, however those approaches tend to rely on hand-defined features and representations. The latter being built on top of various perception algorithms like object detection or semantic segmentation.

Recently with the surge of deep learning for computer vision and reinforcement learning various new methods have been proposed to tackle this problem. End-to-end control learning \cite{chen2018learning, mousavian2019visrep}, hybrid approaches combining traditional planning with RL \cite{chaplot2020object, gupta2017cvpr}, among others have been proposed. Many of these works use reinforcement learning as a basis for their control and tend to use only a limited discrete action space. They focus on policies with simple actions like (left, right and straight) and do not address the problem of continuous robot motion control. 
With the goal of achieving efficient \emph{object goal navigation}, i.e. reaching a defined semantic target, and fully exploit robot kinematics, inspired by \cite{drews2017corl}, we propose a technique that combines model-based continuous control approach with perception module that exploits semantic information and mid-level feature representations. Differently from \cite{ramakrishnan2022poni}, instead of predicting only at the frontiers we provide dense predictions: the latter is a more natural and common representation for downstream tasks such as planning and control.

\begin{figure}[t]
  \centering
  \includegraphics[width=1.0\linewidth]{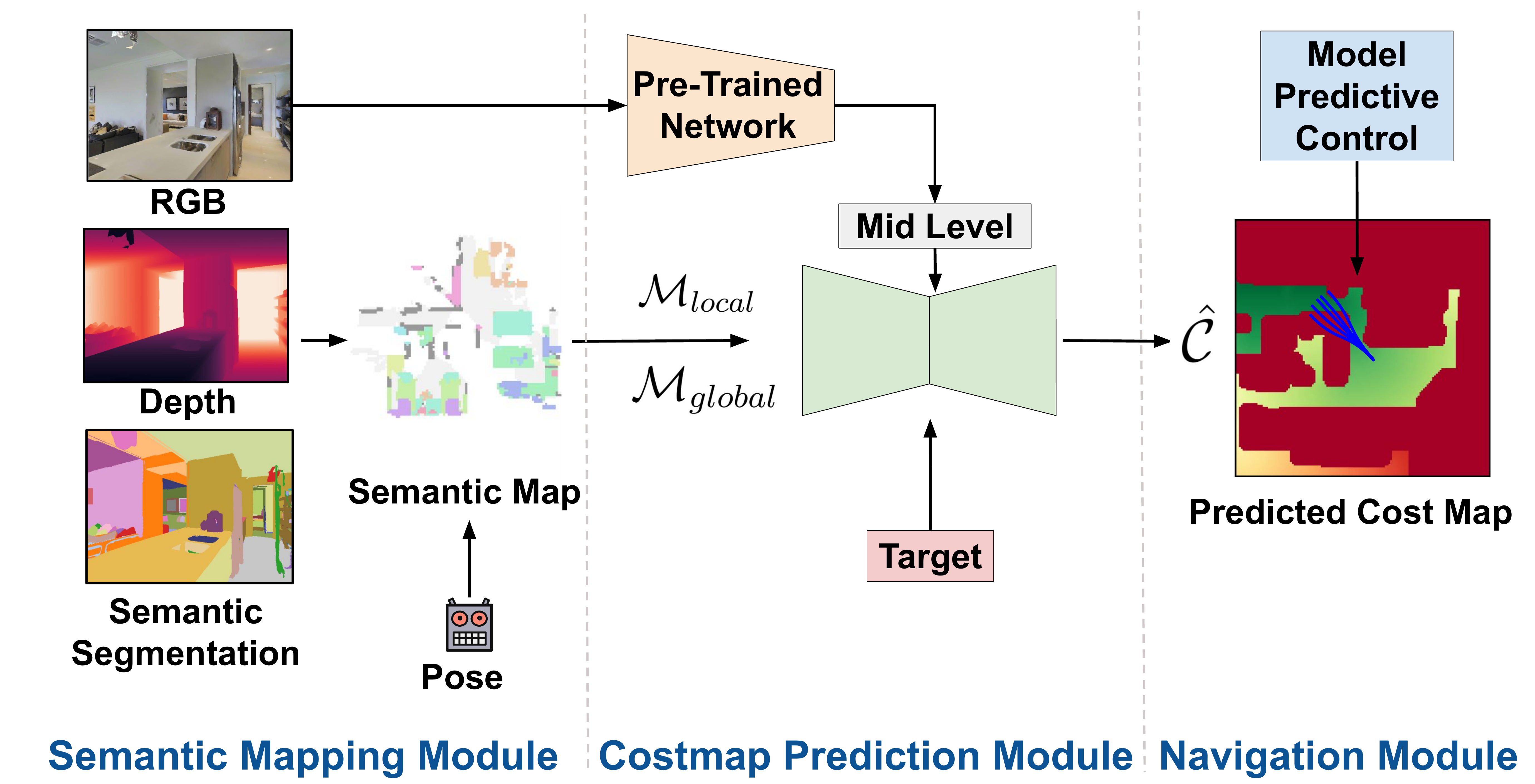}
  \caption{Our framework is composed of three modules: \textit{semantic mapping module} which constructs the map of the environment as the robot explores. The \textit{cost map prediction module} which predicts the cost map for navigation based on the input semantic map, mid-level representation and target object. Finally, the \textit{navigation module} generates the optimal control using a sampling-based MPC.}
  \label{fig:framework}
\end{figure}

\begin{figure*}[t]
  \centering
  \includegraphics[width=0.9\linewidth]{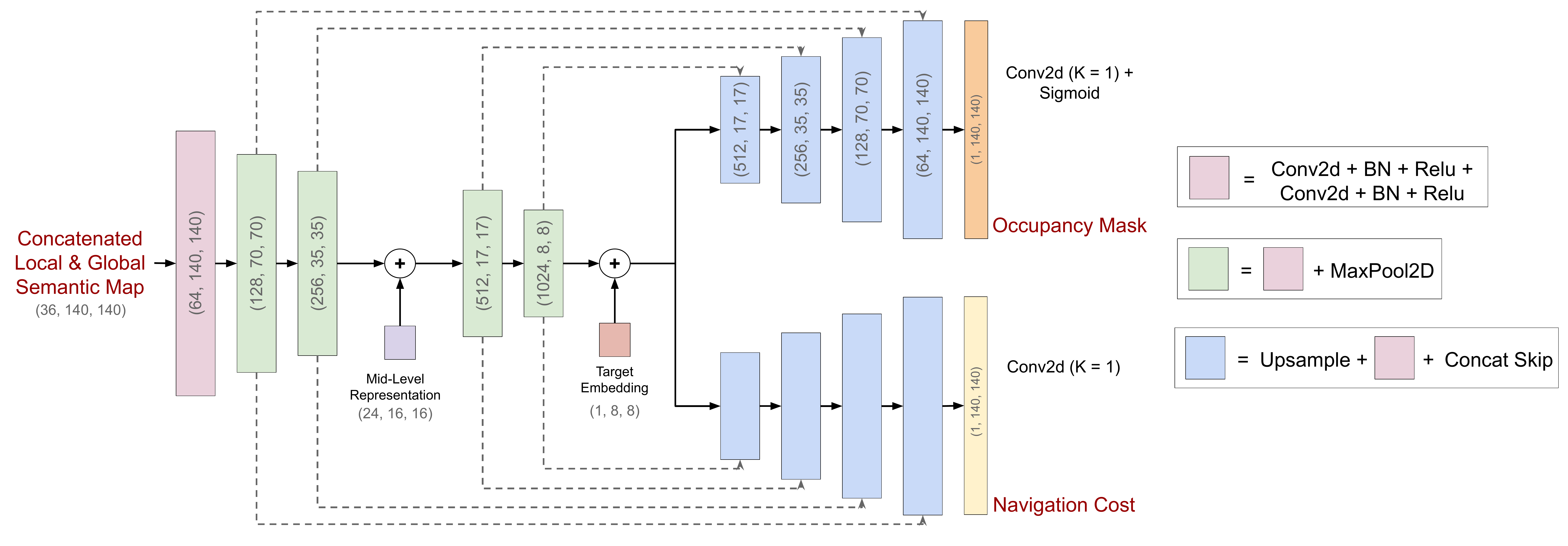}
  \caption{Our detailed network architecture for cost map prediction. Dashed arrows denote skip connections. Kernel size K as 3, stride S as 1 and padding P as 1 are taken unless specified otherwise. The mid-level representations are fused according to orientation bin as shown in \figref{fig:midlevel}. In the end, the occupancy mask and navigation cost are fused to get the final cost map prediction.}
    \label{fig:network}
\end{figure*}

We summarize our main contributions as follows:
    
    \textbf{(i)} We propose a U-Net based architecture for dense cost map prediction under partial observability and design loss function along with dataset for training.
    
    \textbf{(ii)} We explore the use of egocentric mid-level visual representations in the network architecture. We present a novel approach of fusing these features to our network in an robot-orientation-aware way. Our experimental evaluation shows that mid-level representations significantly (by 7 percentage points in success rate) improve the navigation performance.
    
    \textbf{(iii)} We further show that the predicted cost maps can be used for semantic navigation in a loosely coupled scheme with sampling-based model predictive control (MPC) in continuous action space.


\section{Our Approach}
\label{sec:main}

For object goal navigation task, the agent $\mathcal{A}$ is spawned in a random position in an uknown environment and is given a target object category $\mathcal{T}$ that the agent has to reach (e.g. a target category like \emph{bed}). The agent has access to sensor observations ($\mathcal{O}_t$) consisting of ego-centric RGB, depth and semantic segmentation provided from the Habitat simulator~\cite{savva2019habitat}. Apart from this, the robot odometry, $\mathbf{x}_t$ is also available. The model of the robot is a differential drive model which is operated
using velocity control (translational velocity $v_t$ and angular velocity $\omega_t$).

Our approach is shown in the \figref{fig:framework}. The whole framework can be divided into three major components. First component is \emph{semantic mapping}, which is discussed in \secref{sec:semmap_gen}. The output from this module is send to \emph{cost map prediction network}, which is discussed in the \secref{sec:cost mapnetwork}. It takes as input the unexplored and incomplete map around the robot to predict the cost map. The loss is defined in \secref{sec:lossfunction}. We use the predicted cost map in the \emph{navigation} module as detailed in the \secref{sec:navigation}.

\subsection{Semantic Map Generation}
\label{sec:semmap_gen}

We follow the approach of Chaplot \etal \cite{chaplot2020object} to construct the semantic map $\mathcal{M}$ of the environment. Overall, it contains the information of obstacles, explored area and top-down semantic information of each grid cell in the  map. We start by transforming the first person semantic segmentation and depth to top-down semantic maps. We use the ego-centric semantic segmentation from the Habitat simulator~\cite{savva2019habitat} and keep the study of using off-the-shelf pre-trained segmentation network as a future work.

Using known camera intrinsic parameters and depth from Habitat, each pixel in the camera image is projected to 3D space along with their semantic label to get a semantic point cloud. The point cloud is then transformed to the world coordinates by using the robot pose $\mathbf{x}_t \in \mathcal{X}$, $\mathcal{X}$ being the space of all possible robot states. The semantic point cloud is converted to top-down 2D semantic map such that each cell has a semantic label with different probabilities for each class. For the $K$ number of semantic classes we have the semantic map of size $(K, N, N)$ where $N$ is the size of local spatial region that we see in each view. We use 16 semantic classes which is a superset of our target classes to represent our semantic map. We further concatenate this map with obstacle mask and explored mask to finally get map $\mathcal{M}_t$ of size $(C, N, N)$ where $C = 2 + K$ for time $t$. This unit generates a local and a global semantic map (see Sec.~\ref{sec:cost mapnetwork}).

\subsection{Cost Map Prediction}
\label{sec:cost mapnetwork}

We use the local semantic map, global semantic map and the mid-level representations \cite{midLevelRepssax2018} as the input to the network. The network architecture is inspired from the U-Net architecture \cite{ronneberger2015unet}. The whole module architecture can be seen in the \figref{fig:network}. We discuss the various components in this section.

\textbf{Semantic Map.} The local semantic map $\mathcal{M}_{local}$ and the global semantic map $\mathcal{M}_{global}$ generated from the \emph{semantic mapping} module in \secref{sec:semmap_gen} are used as input to the network. They are concatenated across the channel before being given as an input to the network. So, the total map input size is $(2C, H, W)$, which for our case is $(36, 140, 140)$. The global map is reduced to spatial size of local map using average pooling.

\textbf{Mid-Level Representation.} 
Mid-level representations are features generated from encoders which have been trained for different downstream tasks like \emph{semantic segmentation}, \emph{denoising}, \emph{curvatures}, \emph{keypoints}, etc. They have shown to be quite effective in RL setting for various downstream tasks \cite{shen2019situationalfusion, midLevelRepssax2018}: they improve generalizability and also performance of an RL agent. 
In our approach, we adopt mid-level representations and show how they can be beneficial for predicting context-aware dense cost maps.

\begin{figure}[t]
  \centering
  \includegraphics[width=0.7\linewidth]{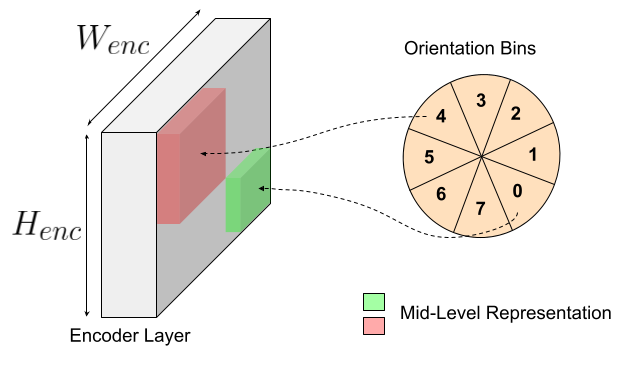}
  \caption{Orientation of the robot decides the orientation bin in which the mid-level feature falls. Based on the bin, the corresponding region (bin region) is set to the mid-level feature while the rest of layer is set to 0 to form the oriented mid-level representation. Each bin is of the size 45$^{\circ}$ so that we have a total of 8 bins to cover all the orientation possibilities. For example, the agent looking towards the north will have the bin 3 and the associated middle top region will have the mid-level feature.}
    \label{fig:midlevel}
\end{figure}

The mid-level representations are extracted from the RGB image using a pre-trained network as used by Sax \etal\cite{midLevelRepssax2018}. We use \emph{semantic segmentation} and \emph{object classification} to help learn about object goal and semantic contexts of the world. Apart from this, we use \emph{depth prediction} since we are learning distance dependent cost map. These are combined at the encoder stage of the network. They are binned according to the orientation of the robot to form \emph{oriented mid-level representation} as explained in \figref{fig:midlevel}. Using this binning technique we make these mid-level representations robot orientation aware and associate with the corresponding semantic input map region.

\textbf{Target Embedding.} We encode the target object category that the agent has to reach. To do this, we create an embedding using the index of the target category. The embedding of size $M$, where $M=64$, is transformed to the spatial size of the encoded latent feature $(H_{enc}, W_{enc})$. It is then concatenated to the latent feature along the channel dimension.

\textbf{Output.} We observed that casting the cost map prediction as a multi-task learning problem leads to better results. Therefore, our network consists of two decoder branches. One predicts the occupancy map $\hat{\mathcal{C}}^{occ}$ and the other decoder branch predicts navigation cost $\hat{\mathcal{C}}^{nav}$. We combine $\hat{\mathcal{C}}^{occ}$ and $\hat{\mathcal{C}}^{nav}$ to get the final cost map prediction $\hat{\mathcal{C}}$. This is done by using an occupancy threshold $\theta_{occ}$ to create a binary occupancy mask from $\hat{\mathcal{C}}^{occ}$, which is then overlayed on the navigation cost $\hat{\mathcal{C}}^{nav}$.

\subsection{Loss Function}
\label{sec:lossfunction}
In this section we describe our multi-term loss function that we formulated to train the network for the cost map prediction. In the following equations we denote the ground truth occupancy mask by $\mathcal{C}^{occ}$ and the ground truth navigation cost by $\mathcal{C}^{nav}$. The dataset generation technique for ground truth has been discussed in the \secref{sec:dataset}.

\textbf{Occupancy Loss.} We use a binary cross entropy loss over the local map region of size $(H,W)$ to learn the occupancy probabilities $\hat{c}^{occ}_{i, j}$ of a map cell $(i, j)$:

\begin{equation}
\mathcal{L}_{occ}=\frac{1}{H W} \sum_{i, j}-c^{occ}_{i, j} \log \left(\hat{c}^{occ}_{i, j}\right)-\left(1-c^{occ}_{i, j}\right) \log \left(1-\hat{c}^{occ}_{i, j}\right),
\end{equation}
where $c^{occ}_{i, j}$ is 1 in case of obstacle and 0 in case of free space. This loss term is used only for the branch predicting the occupancy map.

\textbf{Cost Map Loss.} To learn the cost map prediction, we regress the navigation cost using the L1 norm, averaged over all valid positions (i.e. navigable area):
\begin{equation}
\mathcal{L}_{cost} = \frac{1}{HW}\sum_{i, j} \left\|\left(c^{nav}_{i, j} - \hat{c}^{nav}_{i, j}\right)\left(1-c^{occ}_{i, j}\right)\right\|_{1},
\end{equation}
where $c^{nav}_{i, j}$ is the normalized ground truth navigation cost and $\hat{c}^{nav}_{i, j}$ is the predicted navigation cost.

\textbf{Gradient Direction Loss.} We introduced this term to make the navigation cost smooth and consistent with the local ground truth gradients. Similar to cost map loss, we only calculate this loss over navigable area of the map:
\begin{equation}
\mathcal{L}_{dir} = \frac{1}{HW}\sum_{i, j}\left( 1 - \frac{\mathbf{g}_{i, j}\cdot\mathbf{\hat{g}}_{i, j}}{|\mathbf{g}_{i, j}|\cdot|\mathbf{\hat{g}}_{i, j}|}\right)\left(1-c^{occ}_{i, j}\right),
\end{equation}
where $\mathbf{g}$ is the gradient for ground truth cost map and $\mathbf{\hat{g}}$ is the gradient for predicted cost map,
\begin{align*}
\mathbf{g}_{i, j} &= \left(\frac{\delta{\mathcal{C}^{nav}}}{\delta{x}}, \frac{\delta{\mathcal{C}^{nav}}}{\delta{y}}\right)_{i, j}\\
\hat{\mathbf{g}}_{i, j} &= \left(\frac{\delta{\hat{\mathcal{C}}^{nav}}}{\delta{x}}, \frac{\delta{\hat{\mathcal{C}}^{nav}}}{\delta{y}}\right)_{i, j}.
\end{align*}

In our experiments we observed that the addition of this term leads to significantly smoother cost maps, which is important for the downstream navigation task.

The total loss then becomes a combination of all these losses,
\begin{equation}
\mathcal{L}_{total} = \alpha_{occ}\mathcal{L}_{occ} + \alpha_{cost}\mathcal{L}_{cost} + \alpha_{dir}\mathcal{L}_{dir},
\end{equation}
with the empirically selected weights $\alpha_{occ} = 1.0$, $\alpha_{cost} = 1.5$ and $\alpha_{dir} = 1.0$.

\subsection{MPC based Navigation}
\label{sec:navigation}
We use the sampling-based MPC approach of IT-MPC (Infomation Theoretic Model Predictive Control \cite{williams2017icra}) for the agent to pick the optimal control sequence. We start by having the initial control sequence $U = \{\mathbf{u}_{0}, \mathbf{u}_{1}, \ldots \mathbf{u}_{H-1}\}$ where $H$ is the horizon of the IT-MPC. Each timestep control, $\mathbf{u}_{t} = \{v_{t}, \omega_{t}\}$ where $v_{t}$ and $\omega_{t}$ are the linear and angular velocity respectively for timestep $t$. Each control in the sequence is then perturbed for $K$ samples to generate noisy control, $\Tilde{U}_k = U + \mathcal{E}_k$ where $\mathcal{E}_k = \{\epsilon_{0}, \epsilon_{1}, \ldots \epsilon_{H-1}\}$. Every noise $\epsilon_{t}$ is sampled from a normal distribution $\mathcal{N}(\mu, \sigma)$ using $\mu$ = 0 and $\sigma$ = 0.35.

For each sample $\Tilde{U}_k$, we generate the cost $\mathcal{S}_k$ using the  
predicted cost map
and the control effort.
\begin{equation}
\mathcal{S}_{k} = \sum_{t=0}^{H-1} \left( \hat{\mathcal{C}}_{t}(\mathbf{x}_{t}) + \mathbf{u}_{t}^{T}Q\mathbf{u}_{t}\right),
\end{equation}
where $\mathbf{x}_t = \{x_{t}, y_{t}, \theta_{t}\}$ is the robot pose and $Q \ge 0$ is the control effort matrix. The robot pose is sampled using a constant velocity differential model using the perturbed linear and angular velocity.

The cost is then used to generate weights $w_k$ of the importance sampling step that obtains the optimal control sequence to execute:
$\beta = \min _{k}S\left(\mathcal{E}^{k}\right), \eta = \sum_{k=0}^{K-1} \exp \left(-\frac{1}{\lambda}\left(S\left(\mathcal{E}^{k}\right)-\beta\right)\right), 
    w_{k} = \frac{1}{\eta} \exp \left(-\frac{1}{\lambda}\left(S\left(\mathcal{E}^{k}\right)-\beta\right)\right)$.
Finally, the control sequence $\mathbf{u}_{t}$ is updated using the weights and control noise.
\begin{equation}
\mathbf{u}_{t+1} = \mathbf{u}_{t} + \sum_{k=1}^{K}w_{k}\epsilon_{t}^{k}.
\end{equation}
Hence, we get the updated control and we apply the first control $\mathbf{u}_{0}$ to the agent.

\textbf{Goal Reacher.} During the exploration, if the target object category $\mathcal{T}$ is observed in the local top-down semantic map $\mathcal{M}_{local, t}$ 
then the corresponding map cells define the goal mask $\mathcal{M}_{goal, t}$. To avoid false positives we remove small regions from the mask.
Then using the remaining goal mask $\mathcal{M}_{goal, t}$ and local occupancy map $\mathcal{M}^{occ}_{local, t}$ we generate cost map for navigation using Fast Marching Method ($\mathrm{FMM}$)~\cite{osher1988fmm}: 
\begin{equation}
\mathcal{C}_{t}^{goal} = \mathrm{FMM}\left(\mathcal{M}^{occ}_{local, t}, \mathcal{M}_{goal, t}\right).
\end{equation}

The agent then drives using this cost map. It declares $\mathtt{done}$ when the cost map value is less than or equal to the cost threshold $\theta_{cost}$ i.e. $\mathcal{C}_{t}^{goal} \leq \theta_{cost}$. In our experiments we used $\theta_{cost} = 0.2$. If the goal turns out to be unreachable due to previously unobserved obstacles then our approach leaves the goal reaching mode and resorts to the predicted cost map to continue the exploration.

\section{Experimental Setup}
\label{sec:exp}

We perform experiments in the real-world indoor environments provided by a large-scale RGB-D dataset Matterport3D (MP3D)~\cite{chang2017mp3dn}. We use a physics-enabled 3D simulator Habitat~\cite{savva2019habitat} to navigate the agent in these environments. To train our cost map prediction network we generate a dataset as described in \secref{sec:dataset}. We describe the evaluation setup and the metrics for cost map prediction and navigation in \secref{sec:evalprediction} and \secref{sec:evalnavigation} respectively. Finally, \secref{sec:implementation} provides important implementation details.

\subsection{Cost Map Prediction Dataset}
\label{sec:dataset}

Our interest is in house-like environments containing objects, such as a couch, a bed, a table, a chair, a plant, etc. Therefore, as a first step, we filter out environments from MP3D dataset which do not contain relevant semantic information e.g. large halls or churches. In addition, we omit the houses containing multiple incorrect object labels, which can impair the training process. 
The remaining 48 houses form our dataset, which is divided into the train, validation and test splits consisting of 36, 4 and 8 houses respectively.

For each house, we sample multiple starting points and randomly select a goal from the set of goals we are considering.  For each floor where the robot is spawned, we get the ground truth top-down semantic map as defined in \cite{cartillier2020semanticn} which is then used to generate the goal map based on the target object. Combining this goal map with the occupancy map from the Habitat simulator, we generate the global ground truth cost map of distances. 

We use an IT-MPC based agent with the ground truth cost maps to reach the goal while we collect the dataset. We record samples at every fourth timestep to reduce redundancy. The number of trajectories taken in a house depend on the size of the house to avoid repetition. This was selected manually for each house upon inspection. The complete generated dataset contains 171412, 16209 and 41949 samples in the train, val, and test splits respectively.

Each training sample consist of i.) local semantic map ($\mathcal{M}_{local}$) of size $140\times140$, ii.) global semantic map ($\mathcal{M}_{global}$) of size $420\times420$ to help capture global context, iii.) ground truth cost map ($\mathcal{C}$) composed of distances to the goal using FMM~\cite{osher1988fmm} and occupancy map, and, iv.) ego-centric RGB for computing the mid-level visual representations~\cite{midLevelRepssax2018}. We also save the orientation of the robot along with the image.

\subsection{Evaluation Setup for Cost Map Prediction}
\label{sec:evalprediction}
We evaluate both occupancy and cost map prediction for our approach. Both the predictions are evaluated on the test split of the generated dataset (\secref{sec:dataset}). The occupancy prediction uses classification metrics of mean F1 score (mF1) and mean Intersection over Union (mIOU) averaged over both free space and occupied space classes. We also report mean pixel accuracy (MPA) for occupancy prediction. For navigation cost prediction, we report average Action Prediction (aAP) which determines the accuracy of picking the right local policy normalised by navigable area. aAP$_5$ determines per-pixel accuracy of picking the correct action based on the lowest cost from 4 basic directions and being stationary. Similarly, aAP$_9$ measures the same but with 8 neighbours and the robot position. aAP$_9$ gives us a more accurate resolution as the agent can move in diagonal directions as well.
\subsection{Evaluation Setup for Object Goal Navigation }
\label{sec:evalnavigation}
For navigation performance we ran the agent on different houses from the test split in the Habitat simulator. There are a total of 8 test houses - for each house we sample 40 random starting positions for the robot along with a random target object to reach. We gather various metrics related to success of reaching the target object like {success} of reaching the goal, SPL~\cite{anderson2018eval}, - which weights the success according to the path length determining its efficiency  and DTS~\cite{chaplot2020object} i.e. Distance to Success, which measures how far is the agent from the success distance, which in our experiments was set to 1\,m. We also measure the smoothness of the final robot trajectory using average acceleration and jerk. We run each experiment for 500 timesteps which is equivalent to 50\,s. 
The target object can be selected from any of the target category list that we consider: {\textit{bed}, \textit{chair}, \textit{sink}, \textit{plant} or \textit{couch}}.

\subsection{Implementation Details}
\label{sec:implementation}
The cost map prediction network was implemented in PyTorch. During training, we applied data augmentation by randomly rotating (with a probability of 0.15) the input semantic maps and the target cost maps by 90$^{\circ}$, 180$^{\circ}$ or 270$^{\circ}$.
We used the SGD optimizer and a constant weight decay factor of $0.01$. In all experiments, the learning rate followed the cosine decay schedule with a warmup phase of 25 epochs with a peak and terminal learning rates being 15e-5 and 1e-5 respectively. All cost map prediction models were trained for 200 epochs. For the MPC we used an horizon $H=50$.

\section{Experimental Results}
\subsection{Quantitative Results}

In this section, we analyze the quantitative results of our approach for cost map prediction and object goal navigation. 

\begin{figure}[t]
  \centering
  \begin{minipage}{0.49\linewidth}
  \includegraphics[width=\linewidth]{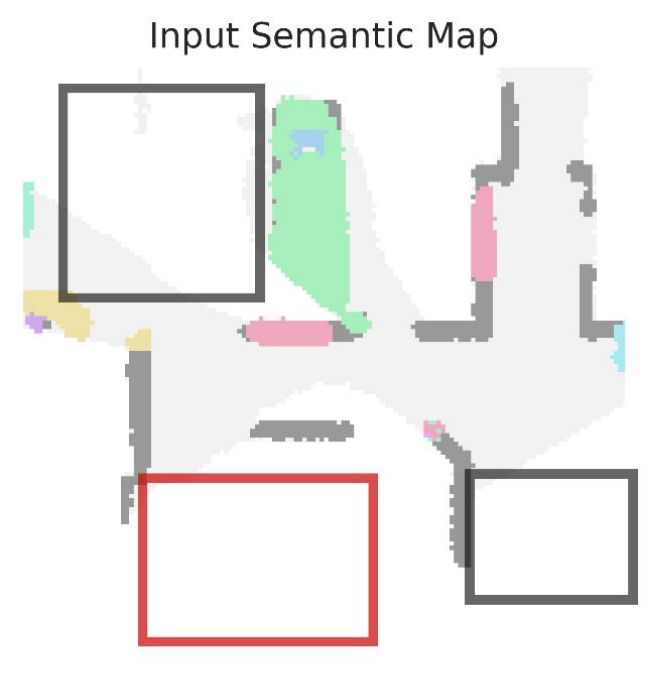}
  \end{minipage}
  \begin{minipage}{0.49\linewidth}
  \includegraphics[width=\linewidth]{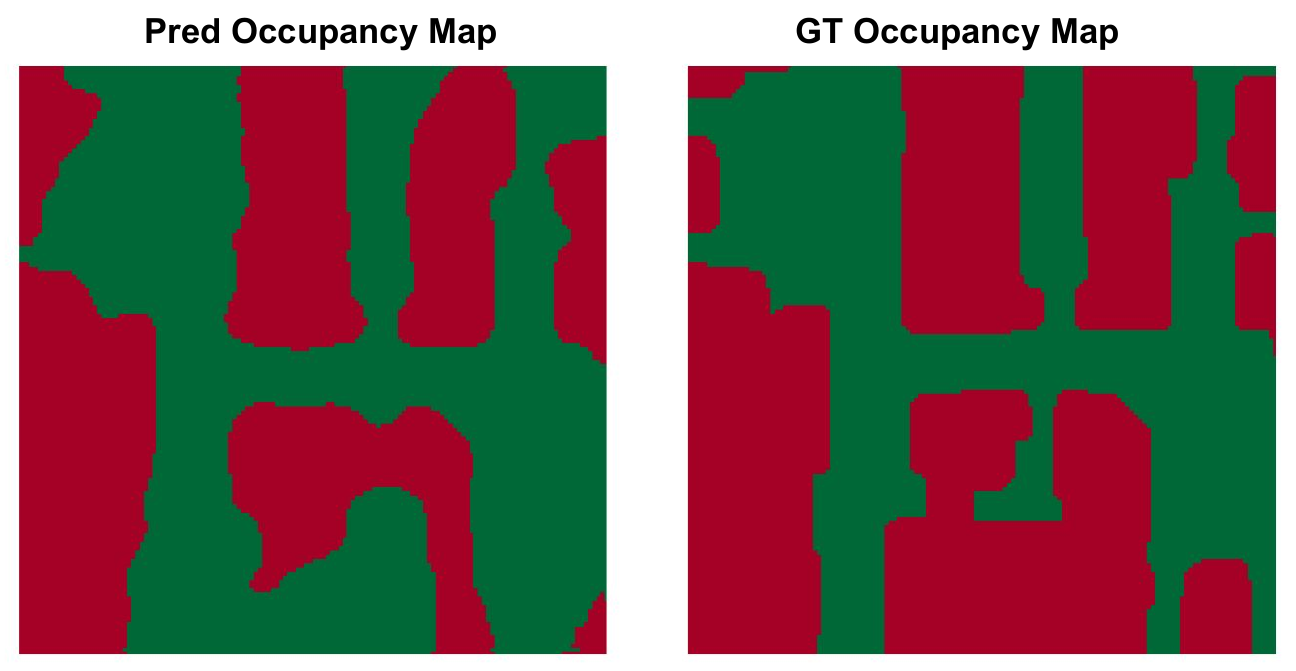}\\
  \includegraphics[width=\linewidth]{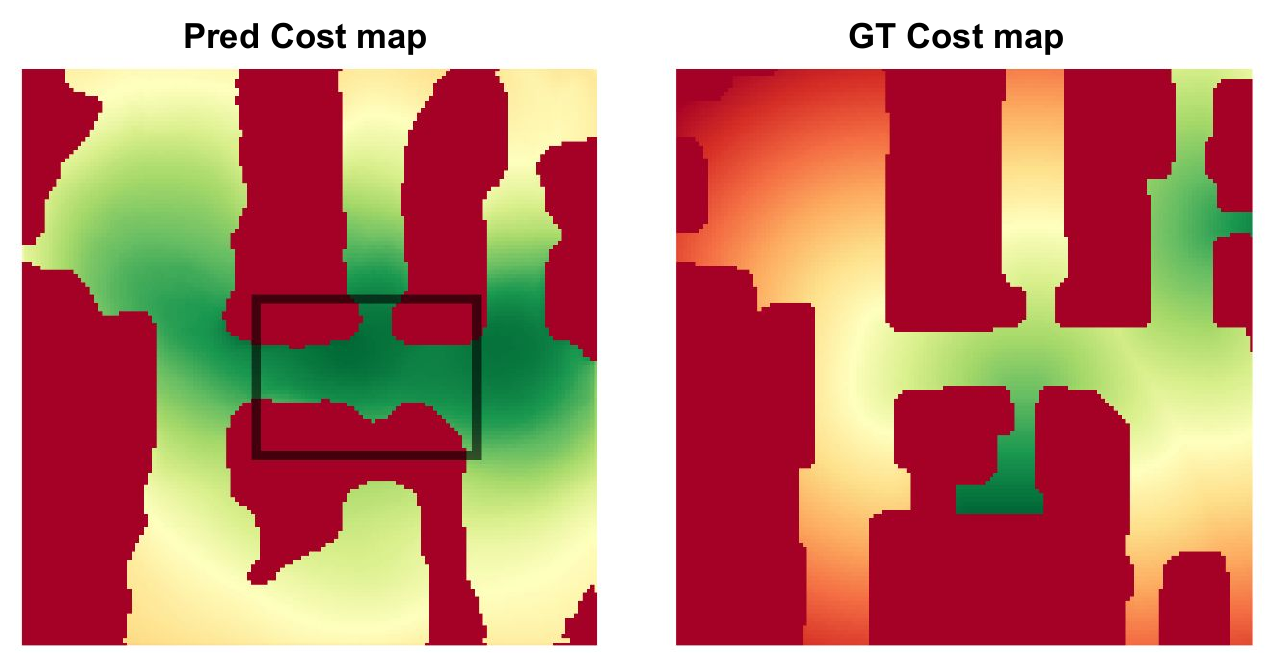}
  \end{minipage}

 \caption{The left image shows the input semantic map where, white is unexplored area, gray is free space and the rest is other semantic regions. The predicted occupancy map on the top right matches well with the ground-truth occupancy map in the black box regions. Whereas for the red box, our method is extrapolating free space. Here red is occupied and green is free space. For the cost maps on the bottom right, the predicted cost is able to capture the relative lower cost in the center of the map.
 For cost map, \textit{red} to \textit{green} represent \textit{high} to \textit{low} cost.}
  \label{fig:predictionexample}
\end{figure}

\textbf{Prediction.} For occupancy mask prediction, we get an MPA of 78.76\%, mF1 of 75.88\% and mIOU of 62.93\%. We observe that the F1 score of occupied region is 80.68\% and that of free region is 71.08\%. This shows that our occupancy mask prediction approach is inclined to predict the occupancy class better than the free space class. Similar trend was seen for IOU. For occupied region, it is 69.06\% and for free region is 56.79\%. We get scores for aAP$_5$ and aAP$_9$ as 37.45\% and 33.39\% respectively. An example of our prediction can be seen in \figref{fig:predictionexample}.

\begin{table*}[h]
\centering
\resizebox{1.9\columnwidth}{!}{%
  \begin{tabular}{l|ccccccc}
    \toprule
     Methods & SR $\uparrow$ & SPL $\uparrow$ & DTS($m$) $\downarrow$ & Time $\downarrow$ &  Acc($ms^{-2}$, $rads^{-2}$) $\downarrow$ & Jerk($ms^{-3}$, $rads^{-3}$) $\downarrow$ \\
      \midrule
    GT cost map & 1.000 & 0.922 & 0.211 & 147.459 & [0.15, 2.041] & [2.371, 35.006]\\
    \midrule
    Privil. Random & 0.250 & 0.208 & 7.214 & 402.795 & [1.607, 8.920] & [28.004, 156.095]\\
    Our Approach & \textbf{0.447} & \textbf{0.329} & \textbf{4.542} & \textbf{327.877} & \textbf{[0.673, 7.318]} & \textbf{[11.788, 128.158]}\\
    \bottomrule
  \end{tabular}
}
\caption{Navigation performance comparison. GT cost map provides an indication of the best possible metrics.}
\label{table:baseline_comparison_GT_success_eps}
\end{table*}

\textbf{Navigation.} We compare our approach for object goal navigation with the following agents:

1. \textit{Priviliged Random Agent}: It picks a random value from the allowable set of linear and angular velocity to use as an action, it is made \emph{privileged} by providing the \emph{goal reacher}, which is semantically informed. This helps us to see the improvement in exploration by our agent.

2. \textit{MPC with GT cost map}: This agent has access to ground truth cost map which is then used for navigation by the IT-MPC.
This agent declares the completion of episode when the goal is visible. This agent is useful to understand the upper bound on the performance.

As can be seen in \tabref{table:baseline_comparison_GT_success_eps}, our agent does not match the performance of the GT agent which is an expected outcome. However, we can clearly see that our agent outperforms the privileged random one. Our approach is able to reach the goal with more success of 0.447 (increase of 0.197) and higher SPL of 0.329 (increased by 0.121) which shows our approach is capable of more efficient exploration. This is further justified by a decrease in DTS of 2.342$m$ - which shows it is able to reach nearer to the goal. Our agent also generates much smoother path as evident from average acceleration and average jerk.

\subsection{Qualitative Results}
\begin{figure}[t]
  \centering
  \includegraphics[width=0.95\linewidth]{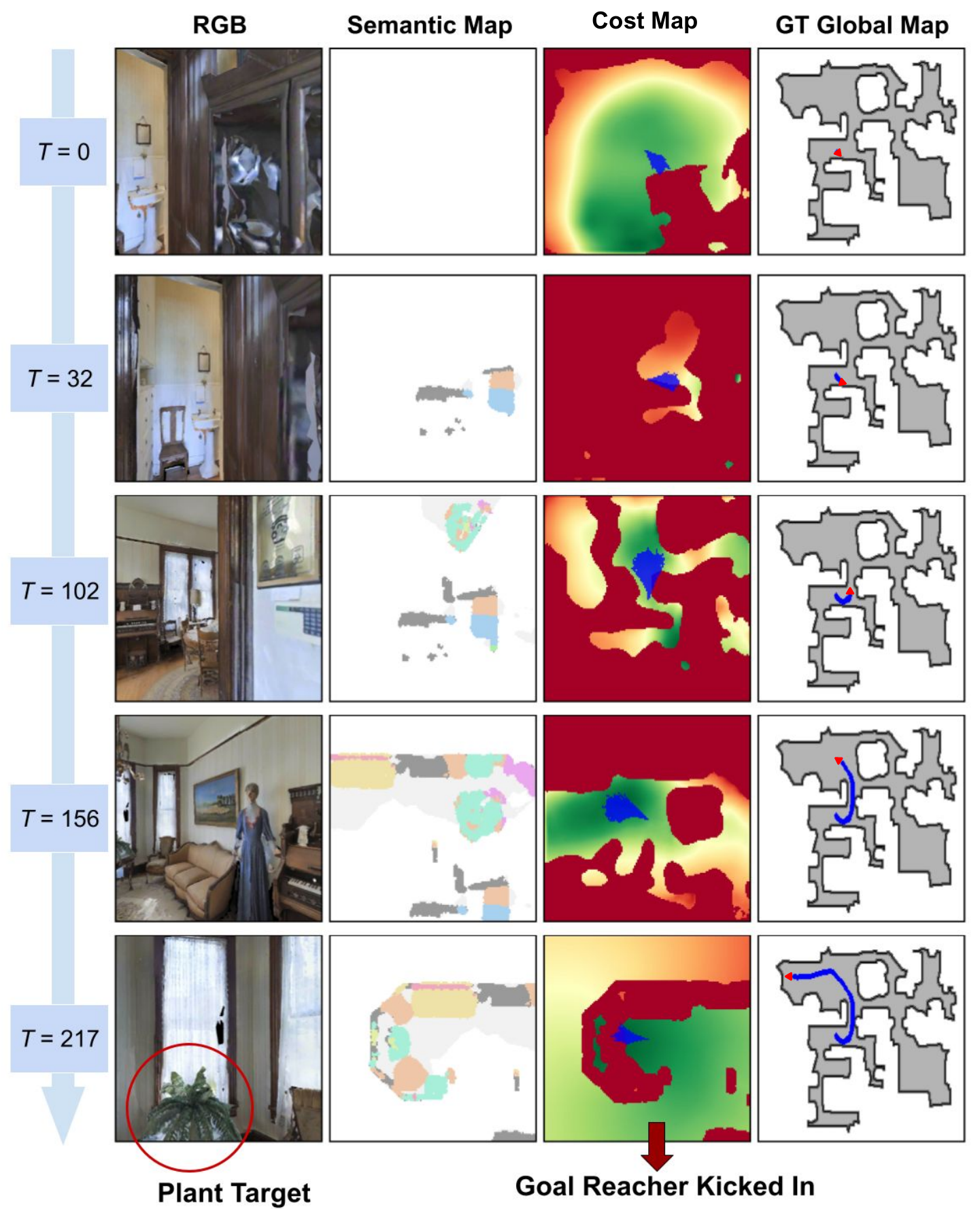}
  \caption{Progression of agent moving in the house over time. The path over time is shown in \textit{blue} in the GT global map with the \textit{red} arrow showing the orientation of the robot. The target goal in this case is \textit{plant} and we see that the agent is able to navigate to the plant efficiently. We also see in the last timesteps the goal reacher is activated. In the cost maps, the \textit{green} regions show low cost and \textit{red} the high-cost regions. Samples from MPC are also shown in the cost maps in \textit{blue}.}
  \label{fig:nav_result_1}
\end{figure}

In this section we present the qualitative results to show how our agent successfully completes the task of object goal navigation on one of the test sequence.
\figref{fig:nav_result_1} shows an example of a successful trajectory taken by our agent in reaching the target object. We can see how the agent progresses towards the goal as it builds the semantic map of the environment. The predicted cost map efficiently guides the agent near the goal area. At timestep $217$ the goal reacher (described in \secref{sec:navigation}) gets activated as the target object (i.e. \textit{plant}) becomes visible in the local semantic map. Finally, the cost map generated by the goal reacher drives the agent to the goal.

\subsection{Ablation Study}
\begin{table*}[h]
\centering
\resizebox{1.9\columnwidth}{!}{%
  \begin{tabular}{l|ccc|cccccc}
    \toprule
     \multirow{2}{*}{Methods} &
      \multicolumn{3}{c}{\textit{Cost Map Prediction}} &
      \multicolumn{5}{c}{\textit{Object Goal Navigation}} \\
     & MPA(\%) $\uparrow$ & mF1(\%) $\uparrow$ & mIOU(\%) $\uparrow$ & aAP$_5$(\%) $\uparrow$ & aAP$_9$(\%) $\uparrow$ & SR $\uparrow$ & SPL $\uparrow$ & DTS($m$) $\downarrow$\\
      \midrule
      Only Semantic Map & \textbf{79.17} & \textbf{76.32} & 63.34 & 36.49 & 32.47 & 0.377 & 0.290 & 4.803\\
      Only Mid-Level  & 79.11 & \textbf{76.32} & \textbf{63.47} & 37.21 & 33.11 & 0.402 & 0.287 & 4.716\\
      Both (Our Approach) & 78.76 & 75.89 & 62.93 & \textbf{37.45} & \textbf{33.39} & \textbf{0.447} & \textbf{0.329} & \textbf{4.542}\\
    \bottomrule
  \end{tabular}
 }
\caption{Performance comparison for different ablations.}
\label{table:mid_level_compare_prediction}
\end{table*}

In this ablation study, we compare how the different input information affect our results. We compare three variants: \emph{Only Semantic Map}, which has only semantic map as the input to the cost map prediction module, \emph{Only Mid-Level}, which has input map without semantics along with the mid-level representations and finally, \emph{Our Approach}, with both semantic map and mid-level input.  
Results in \tabref{table:mid_level_compare_prediction} show that our approach achieves better results. 

The \emph{Only Mid-Level} already reaches a success rate of 0.402 while the one with \emph{Only Semantic Map} has a success rate of 0.377. Addition of mid-level representations greatly improves the success rate of the \emph{Only Semantic Map} by 7 percentage points. While addition of semantic map to \emph{Only Mid-Level} achieves an improvement of 4.5 percentage points. This clearly shows that mid-level representations are beneficial for achieving better object goal navigation efficiency. A combination of both semantic representations leads to the best result in terms of success, SPL and DTS.
\section{Conclusion}
\label{sec:conclusion}
In this work, we present an approach to predict dense and context-aware cost maps for object goal navigation. Our proposed network architecture includes a novel way of fusing mid-level representations which takes into account the orientation of the robot. We demonstrate that the predicted cost maps can be used by sampling based MPC for semantic robot navigation. Moreover, the experiments indicate that the fusion of mid-level representations brings substantial improvement to the navigation performance. As future work, to better understand the potential of our approach, we will investigate additional semantic navigation baselines in continuous control settings. In addition, we wish to perform experiments on real robots and tackle uncertainties and noise as dense cost maps provide an appropriate base for that.
\bibliographystyle{plain_abbrv}
\bibliography{glorified,new}
\end{document}